# Online Path Planning for AUV Rendezvous in Dynamic Cluttered Undersea Environment Using Evolutionary Algorithms

S.M. Zadeh· A.M. Yazdani· K. Sammut· D.M.W Powers

Centre for Maritime Engineering, Control and Imaging,
School of Computer Science, Engineering and Mathematics, Flinders University, Adelaide, SA 5042, Australia

somaiyeh.mahmoudzadeh@flinders.edu.au
amirmehdi.yazdani@flinders.edu.au
karl.sammut@flinders.edu.au
david.powers@flinders.edu.au

**Abstract-** In this study, a single autonomous underwater vehicle (AUV) aims to rendezvous with a submerged leader recovery vehicle through a cluttered and variable operating field. The rendezvous problem is transformed into a Nonlinear Optimal Control Problem (NOCP) and then numerical solutions are provided. A penalty function method is utilized to combine the boundary conditions, vehicular and environmental constraints with the performance index that is final rendezvous time. Four evolutionary based path planning methods namely Particle Swarm Optimization (PSO), Biogeography-Based Optimization (BBO), Differential Evolution (DE), and Firefly Algorithm (FA) are employed to establish a reactive planner module and provide a numerical solution for the proposed NOCP. The objective is to synthesize and analyze the performance and capability of the mentioned methods for guiding an AUV from an initial loitering point toward the rendezvous through a comprehensive simulation study. The proposed planner module entails a heuristic for refining the path considering situational awareness of environment, encompassing static and dynamic obstacles within a spatiotemporal current fields. The planner thus needs to accommodate the unforeseen changes in the operating field such as emergence of unpredicted obstacles or variability of current field and turbulent regions. The simulation results demonstrate the inherent robustness and efficiency of the proposed planner for enhancing a vehicle's autonomy so as to enable it to reach the desired rendezvous. The advantages and shortcoming of all utilized methods are also presented based on the obtained results.

## 1 Introduction

AUV rendezvous with a mother ship, mobile recovery station or submerged vehicle has been an area of interest in recent underwater surveys and oceanic exploration [1-3]. It provides a facility for updating the mission, refueling, data transferring and results in increasing AUV's endurance and extension of underwater operation. To do so, the vehicle needs to have a certain level of autonomy in the path planning procedure. The path planner should be capable of generating a trajectory for at the vehicle's heading and velocity profile to safely guide the vehicle form loitering pint to the desired rendezvous area, while optimizing a performance index such as flight time or energy expenditure. However, the AUV's operation through a non-characterized and cluttered undersea environment adds some difficulties for reliable and optimal rendezvous process [4] On one hand, existence of moving obstacles may lead to change the rendezvous conditions, as time goes on, and on the other hand, variability of ocean current components has considerable impact in drifting the vehicle from target of interest [5]. In such a situation, adaptability of the path planner to environmental variabilities is a key element to carry out the mission safely and successfully. An efficient trajectory produced by the path planner enables an AUV to cope with adverse currents as well as exploit desirable currents to enhance the operation speed that results in considerable energy saving.

The first step toward achieving a satisfactory rendezvous is to find a suitable and efficient method of path planning. In general, the challenges associated with path planning can be considered from two perspective; firstly, the mechanism of the algorithms being utilized, and secondly the competency of the techniques for real-time applications. A plethora of research has been conducted based on deterministic methods for solving unmanned vehicle path planning [6-9]. Path planning based on deterministic methods is carried out by repeating a set of predefined steps that search for the best fitting solution to the objectives [6]. For instance, a non-linear least squares optimization technique is employed for AUV path planning [7] and a sliding wave front expansion algorithm is applied to generate an appropriate path for AUVs in the presence of strong current fields [8]. In [9], level set methods are exploited to provide an energy-efficient optimal path for an AUV considering the benefit of current vectors. All above mentioned deterministic methods demonstrate weak real-time performance and expensive computational cost in high-dimensional problems.

For a group of problems in which the deterministic techniques and classic search methods are not capable of finding exact solutions, the heuristic-grid search approaches are good alternatives with fast computational speed, specifically in dealing with multi-objective optimization problems. Some heuristic search methods such as Dijkstra's, A*, and D* algorithms have been provided and applied to the AUV path planning problem in recent years [10-12]. In [13,14] A* method is utilized to provide a time optimal path with minimum risk for a glider path planning according to an offline data set. D* algorithm operates based on a linear interpolation-based strategy that allows frequently update of heading directions; however, it is computationally expensive in high-dimensional problems [12].

The Fast Marching (FM) algorithm is another approach proposed to solve the AUV path planning problem that uses a first order numerical approximation of the nonlinear Eikonal equation. This method is accurate but also computationally expensive than A*[15]. The upgraded version of FM known as FM* or heuristically guided FM is employed for AUV path planning problem [16]. The FM* keeps the efficiency of the FM along with the accuracy of the A* algorithm, but it is restricted to use linear anisotropic cost to attain computational efficiency. Generally, the heuristic grid-search based methods are criticized because of their discrete state transitions, which restrict the vehicle's motion to discrete set of directions. Particularly, the main drawback of these methods is that their time complexity increases exponentially with increasing problem space, which is inappropriate for real-time applications. This is the case also with the mixed integer linear programming (MILP) method that is used for handling the multiple AUV path planning problem in a variable environment [17].

Meta-heuristic optimization algorithms are efficient methods in dealing with path planning problems of an NP-hard nature. For the AUV path planning problem in a large-scale operation area, satisfying time and collision constraints is more important than providing exact optimal solution; hence, having a quick acceptable path that satisfies all constraints is appropriate than taking a long computational time to find the best path. Meta-heuristic evolution-based optimization algorithms are capable of being implemented on a parallel machine with multiple processors, which can speed up the computation process; hence, these methods are fast enough to satisfy time restrictions of the real-time applications [18,19]. The evolution-based strategies, on the other hand, propose robust solutions that usually correspond to the quasi-optimal solutions [20].

A group of meta-heuristic evolution-based search algorithms has been applied for the path planning problem. In [21], Genetic Algorithm (GA) is utilized to determine an energy efficient path for an AUV encountering a strong time/space varying ocean current field. In [22], an energy efficient path considering time-varying ocean currents is generated by means of a Particle Swarm Optimization (PSO) algorithm. A Differential Evolution (DE) based path planner is applied on an underwater glider traveling in a dynamic ocean structure [23,24] and an off-line path planning method based on Quantum-based PSO (QPSO) is offered for path planning of an unmanned aerial vehicle[25]. The online version of the QPSO-based path planner for dealing with dynamic environment is proposed in [26,27].

Although various evolution based path planning techniques have been suggested for autonomous vehicles, AUV-oriented applications still face several difficulties when operating across a large-scale geographical area. The recent investigations on path planning encountering variability of the environment have assumed that planning is carried out with perfect knowledge of probable future changes of the environment [28,29], while in reality environmental events such as currents prediction or obstacles state variations are usually difficult to predict accurately. Even though available ocean predictive approaches operate reasonably well in small scales and over short time periods, they produce insufficient accuracy to current disposition prediction over long time periods in larger scales, specifically in cases with lower information resolution [30]. The computational complexity grows exponentially with enlargement of search space dimensions. Moreover, current variations over time can change the location and pose of the suspended obstacles and dock station in the case of using mobile docks. Consequently unexpected obstacles may drift across the planned trajectory. Therefore, proper estimation of the behavior of a large uncertain dynamic environment outside of a vehicle's sensor coverage range, is impractical and unreliable, so any pre-planned trajectory based on fixed maps may change to be invalid or inefficient. This becomes even more challenging in larger dimensions, where a huge load of data about update of whole environment condition should be computed repeatedly. This huge data load from whole environment should be analyzed continuously every time that path replanting is required, which is computationally inefficient and unnecessary as only awareness of environment in vicinity of the vehicle such that the vehicle can be able to perform reaction to environmental changes, is enough.

Considering the difficulties associated with operating in large scale underwater fields, the need for an efficient and reliable path planning for having a successful rendezvous mission is clear. Therefore, the main contribution of this paper is threefold. First, the rendezvous problem at hand is represented as a Nonlinear Optimal Control Problem (NOCP) by defining the loitering and rendezvous point as boundary conditions; static and moving obstacles as path constraints; bounds on the vehicle states associated with the limits of vehicle actuators; and Mayer cost function as a final rendezvous time. Thereafter, using penalty function method, an augmented cost function embodied all constraints is made. Second, we mathematically formulate the realistic scenarios that AUV usually is dealing with in a large scale undersea environment. This provides with use of a real map containing current and obstacles information; furthermore, the uncertainty of environment and navigational sensor suites are carefully taken into account by simulating imperfectness of sensory information in dealing with different types of moving objects. Third, is to utilize four evolutionary methods called PSO, DE, BBO and FA to provide a numerical solution for the proposed NOCP. Having significant flexibility for approximating complex trajectories, B-Spline curves are utilized to parameterize the desired path. The combination of meta-heuristic-B-Spline technique provides an effective tool to solve the AUV's rendezvous problem. Considering the results of simulation, performance comparison between all applied methods is undertaken.

The paper is organized as follows. The underwater rendezvous scenario and mathematical model of the vehicle is presented in Section 2. In Section 3, the optimal control framework for the rendezvous problem together with the mechanism of the online path planner is explained. Modelling of underlying operating environment including the obstacles and current vector field is described in Section 4. An overview of the utilized evolutionary methods and their implementation for the path planning problem is discussed in Section 5. The discussion on simulation results are provided in Section 6. The Section 7 presents the conclusion for the paper.

## 2 Rendezvous Scenario and Mathematical Model of the Vehicle

The scenario employed in this paper is for the AUV to rendezvous with a mobile recovery system, which is another larger AUV. We use the terminologies of follower and leader for the AUV waiting in loitering and rendezvous areas, respectively. The leader AUV follows a rectangular rendezvous pattern (Fig.1) and continuously is broadcasting the final rendezvous information. This piece of information encompasses position, course, depth, and rendezvous time. The follower AUV is following a circular loitering pattern (Fig.1) and waiting to receive a first rendezvous message form the leader one. Once the first message is received, the follower starts to compute the path considering all vehicular and environmental constraints. In this situation, if there exists a feasible solution, a 'proceed' message is transmitted and the proposed path and corresponding trajectory are used for rendezvous purpose.

In the worst case, when a feasible path is not achievable, a 'cancel' message is transmitted from the follower to the leader and the follower waits for the next rendezvous message. In this study, it is assumed that there is a perfect link of communication between the leader and follower AUVs. The main objective for the rendezvous problem in summary is to compute an optimal path for the follower from its current position to the final rendezvous point in the pre-set travel time while obeying all constraints. This forms a NOCP.

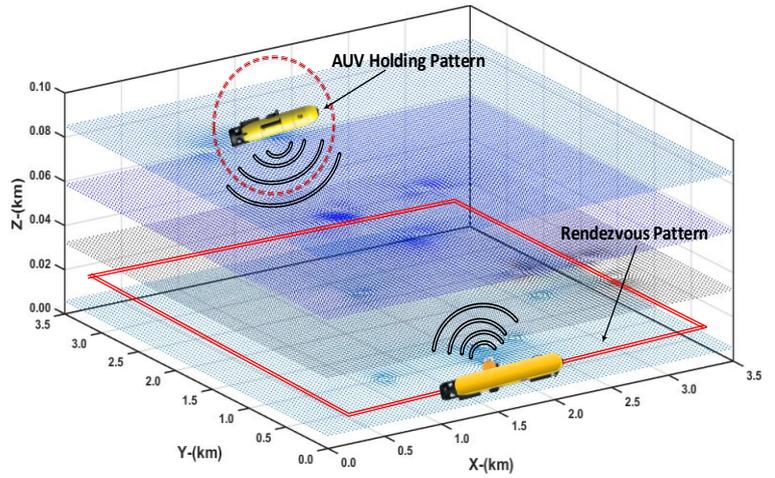

**Fig.1.** Rendezvous scenario

To sketch the corresponding NOCP, first the AUV model is described. The six degree of freedom (DoF) model of an AUV comprise translational and rotational motion. Typically, two frames called North-East-Down (NED) represented by $\{n\}$ and body ($\{b\}$) are used to present the equation of motion [31]. The state variables corresponding to $\{n\}$ and $\{b\}$ frame are shown in Eq.(1)-(2) respectively:

$$\eta : (x, y, z, \phi, \theta, \psi) \quad (1)$$
$$\upsilon : (u, v, w, p, q, r) \quad (2)$$

where $x, y, z$ are the position of vehicle with respect to the North, East, Down, in $\{n\}$-frame, respectively; and $\phi, \theta, \psi$ are the Euler angles called roll, pitch, and yaw, respectively; $u, v, w$ are surge, sway and heave components of water referenced velocity ($\upsilon$) in the $\{b\}$-frame, respectively; and finally $p, q, r$ are roll rate, pitch rate and yaw rate in the $\{b\}$-frame, respectively.

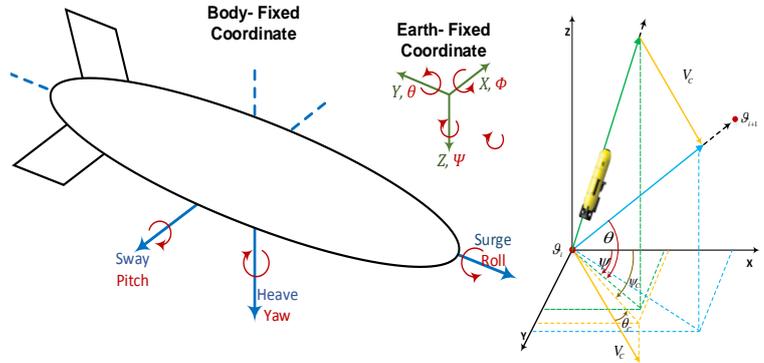

**Fig.2.** Vehicle's and ocean current coordinates in NED and body frame

The kinematics of the AUV can be described by a set of ordinary differential equation shown in Eq.(3), in which the heading and flight path angles are calculated as Eq.(4)-(5), respectively.

$$\begin{aligned}\dot{x} &= u\cos\psi\cos\theta - v\sin\psi + w\cos\psi\sin\theta \\ \dot{y} &= u\sin\psi\cos\theta + v\cos\psi + w\sin\psi\sin\theta \\ \dot{z} &= -u\sin\theta + w\cos\theta\end{aligned} \quad (3)$$

$$\begin{aligned}\psi &= \tan^{-1}\left(\frac{\Delta y_\vartheta}{\Delta x_\vartheta}\right) = \tan^{-1}\left(\frac{|y_{i+1} - y_i|}{|x_{i+1} - x_i|}\right) \\ \theta &= \tan^{-1}\left(\frac{-\Delta z_\vartheta}{\sqrt{\Delta x_\vartheta^2 + \Delta y_\vartheta^2}}\right) = \tan^{-1}\left(\frac{-|z_{i+1} - z_i|}{\sqrt{(x_{i+1} - x_i)^2 + (y_{i+1} - y_i)^2}}\right)\end{aligned} \quad (4)$$

$$r = \dot{\psi} = \left(\frac{\dot{\psi}_i - \dot{\psi}_{i-1}}{h}\right) \tag{5}$$

The proposed path planner in this study, generates the potential path $\wp_i$ using B-Spline curves. It means that, parameterization of the path coordinates in the problem space is established upon the B-Spline curves. The curve is captured from a set of control points like $\vartheta = \{\vartheta_1, \vartheta_2, ..., \vartheta_i, ..., \vartheta_n\}$ in the problem space. These control points play a substantial role in determining the optimal path and basically the optimization process is summarized into finding the best 3D-position for the control points in the problem space. More details are given in section 5. Considering Eq.(4)-(5), the calculated velocity $\upsilon$ should be oriented along the segment $\vartheta_i(x_i,y_i,z_i)$ and $\vartheta_{i+1}(x_{i+1},y_{i+1},z_{i+1})$ according to vehicle's motion direction. Considering Fig.2, the component of current vector $(u_c,v_c,w_c)$ in the $\{n\}$-frame is defined by Eq.(6):

$$\begin{aligned} u_C &= |V_C|\cos\theta_C \cos\psi_C \\ v_C &= |V_C|\cos\theta_C \sin\psi_C \\ w_C &= |V_C|\sin\theta_C \end{aligned} \tag{6}$$

where $V_C$ is magnitude of current velocity, $\psi_c$ and $\theta_c$ are directions of current in horizontal and vertical planes, respectively. It is assumed the vehicle travels with a constant thrust power or in other words has a constant water-referenced velocity $\upsilon$. Applying the water current components $(u_C,v_C,w_C)$ of the ocean environment, the surge $(u)$, sway$(v)$, and have$(w)$ components of the vehicle's velocity are calculated by Eq.(7).

$$\begin{aligned} u &= |\upsilon|\cos\theta\cos\psi + |V_C|\cos\theta_C \cos\psi_C \\ v &= |\upsilon|\cos\theta\sin\psi + |V_C|\cos\theta_C \sin\psi_C \\ w &= |\upsilon|\sin\theta + |V_C|\sin\theta_C \end{aligned} \tag{7}$$

## 3 Rendezvous Problem Formulation Using NOCP Framework and Online Path Planning Strategy
### 3.1 NOCP Framework

In this part, the proposed rendezvous problem is transformed into an optimal control framework as described in the following. Specifically, the goal is to find a path, containing a state vector $X=[x,y,z,\theta,\psi,u,v,w]^T$ that minimizes the performance index (PI) defined as Eq.(8):

$$PI = (t_f - T_r)^2 \tag{8}$$

where $t_f$ is the computed time of the AUV's rendezvous maneuver, $T_r$ is a pre-set rendezvous time. The problem is subject to the vehicle kinematics, represented by Eq.(3), boundary conditions shown in Eq.(9), path constraint due to the obstacles expressed in Eq.(10), bounds over the states shown in Eq.(11), and finally, a pre-set threshold associated with rendezvous time defined in Eq.(12).

$$\begin{aligned} X(t_0) &= X_0 \\ X(t_f) &= X_f \end{aligned} \tag{9}$$

$$\{x,y,z\} \cap \sum\nolimits_{M,\Theta} = 0 \tag{10}$$

$$\begin{aligned} |u(t)| &\leq u_{\max}; \quad |\theta(t)| \leq \theta_{\max} \\ |v(t)| &\leq v_{\max}; \quad |r(t)| \leq r_{\max} \end{aligned} \tag{11}$$

$$|t_f - T_r| < \varepsilon \tag{12}$$

where $t_0$, indicates the starting mission time, and $X_0$, $X_f$ are initial and final boundary conditions over the states, respectively; $\sum_{M,\Theta}$ describes the coastal boundaries and obstacles (as discussed in Section 4), and therefore, the 3D-path should not have an intersection with these forbidden regions; $\varepsilon$ is a threshold corresponding to the optimality selection of generated path. The first path that satisfies all constraints Eq.(9)-(12) is selected as an suitable solution.

Due to its abstract principle and ease of implementation, the penalty function method is a useful tool to use with constraints optimization problems [32]. Using this principle, a new objective function is formed that is the sum of the performance index mentioned in Eq.(8) and a penalty function $F_p(X)$ depending on the constraint violations. By doing this, the proposed NOCP is represented as an unconstrained problem. This augmented from is shown in Eq.(13) in which $\beta_i$ is a weighting factor that corresponds to Eq.(9)-(12). In the sequel, the scaled augmented objective function is represented by Eq.(14).

$$J = PI + \beta_i . F_p(X) \tag{13}$$

$$\begin{aligned} J_{scaled} &= \frac{(t_f - T_r)^2}{T_r^2} + \beta_1 \left(\frac{\max(0,|u| - u_{\max})^2}{u_{\max}^2}\right) + \beta_2 \left(\frac{\max(0,|v| - v_{\max})^2}{v_{\max}^2}\right) + \beta_3 \left(\frac{\max(0,|\theta| - \theta_{\max})^2}{\theta_{\max}^2}\right) + \\ &\quad \beta_4 \left(\frac{\max(0,|r| - r_{\max})^2}{\theta_{\max}^2}\right) + \beta_5 \left(\frac{\max(0,|t_f - T_t| - \varepsilon)^2}{\varepsilon^2}\right) + \beta_6 \left(\frac{\min(0, D_{\wp_{x,y,z}} - \varepsilon_{\sum_{M,\Theta}})^2}{\varepsilon_{\sum_{M,\Theta}}^2}\right) + \beta_7 \left(\frac{(X(T_r) - X_f)^2}{X_f^2}\right) \end{aligned} \tag{14}$$

where $D_{\wp x,y,z}$ represents distance from any point in a 3D-coordinate along the path to the threshold associates with coastal area in the map and obstacles, defined by $\varepsilon_{\sum M,\Theta}$. The augmented objective function shown in Eq.(14), takes into account physical limitations of the vehicle (represented by the constraints over the states), satisfaction of boundary conditions and collision avoidance (represented by the 7$^{th}$ term in Eq.(14)) while minimizing the performance index.

### 3.2 Online Path Planning Strategy

The underwater variable environment poses several challenges for AUV deployment. For instance, variable ocean currents, no-fly zones, and moving obstacles make the AUV mission problematic and cumbersome. One solution to overcome the mentioned difficulties is to establish an on-line path planning approach that incorporates dynamic and reactive behavior. Through this approach, the vehicle exercises situational awareness of the environment and refines the path until vehicle reaches the specified target waypoint. By doing so, for example, in a large-scale operating field, the vehicle is able to make use of desirable current flows at any time that current map gets updated and as a result use the favourable flows for speeding up its motion and saving the energy usage.

In this paper the on-line path planning mechanism contributes in a form that when the corresponding re-planning flag is triggered the new optimal path is generated from current position to the destination by refining the previous solution. In other words, the solution of previously optimal path is used as an initial solution for computing the new path and current states of the vehicle are replaced as the new initial boundary conditions. By following this approach, first of all, the vehicle is able to cope with the uncertainties of the operating field and in fact this strategy provides a near closed-loop guidance configuration; that can be supported with a minimal computational burden, as there is no need to compute the path from the scratch, as opposed to reactive planning strategy [27]. The online path planning mechanism is summarized in Fig.3.

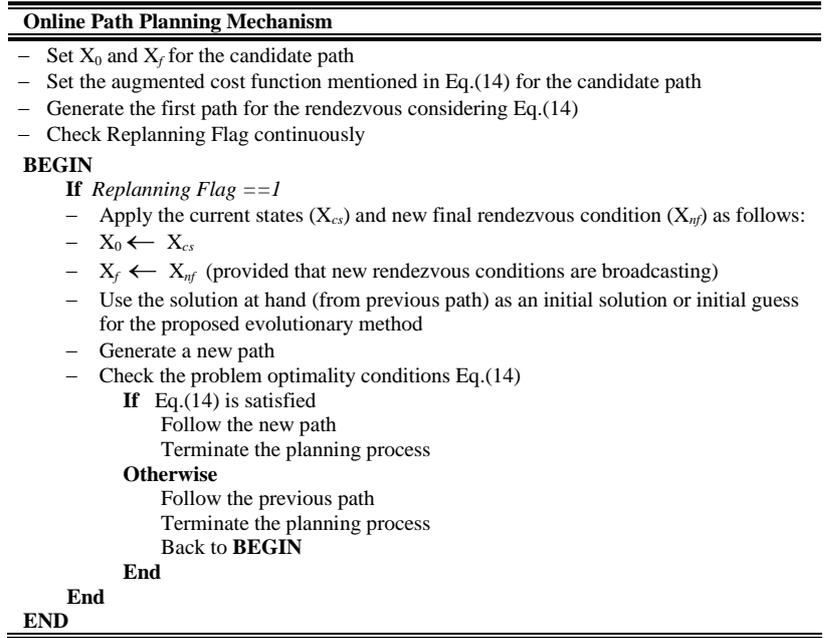

**Online Path Planning Mechanism**
- Set $X_0$ and $X_f$ for the candidate path
- Set the augmented cost function mentioned in Eq.(14) for the candidate path
- Generate the first path for the rendezvous considering Eq.(14)
- Check Replanning Flag continuously

**BEGIN**
    **If** *Replanning Flag ==1*
- Apply the current states ($X_{cs}$) and new final rendezvous condition ($X_{nf}$) as follows:
- $X_0 \leftarrow X_{cs}$
- $X_f \leftarrow X_{nf}$ (provided that new rendezvous conditions are broadcasting)
- Use the solution at hand (from previous path) as an initial solution or initial guess for the proposed evolutionary method
- Generate a new path
- Check the problem optimality conditions Eq.(14)
        **If** Eq.(14) is satisfied
            Follow the new path
            Terminate the planning process
        **Otherwise**
            Follow the previous path
            Terminate the planning process
            Back to **BEGIN**
        **End**
    **End**
**END**

**Fig.3.** Online path planning pseudo code

### 4 Modelling Operational Ocean Environment

Realistic modeling of the operational field including offline map, dynamic/static current flow, moving and static uncertain obstacles provides a thorough test bed for evaluating the path planner in situations that closely match with the real underwater environment. The ocean environment in this study is modelled as a three dimensional environment $\Gamma_{3D}$:{3.5-by-3.5 $km^2$ (x-y), 1000 $m(z)$} embodied uncertain static and moving obstacles in presence of ocean variable current that make the simulation process more challenging and realistic. The path planner designed in this paper is capable of extracting authorized areas of operation through the real map to prevent colliding with coastal areas, islands, and static/dynamic obstacles. In addition, it is capable of compensating for adverse current flows or takes advantage of favorable ones. In the following, the proposed operating field is described in detail.

### 4.1 Offline Map

To model a realistic marine environment, a sample of a real map is utilized as shown in Fig.4. To cluster the coastal areas, and authorized water zones (allowed for deployment) into separate regions, k-means clustering method [33] is employed. It is usually applied for partitioning a data set into k groups. The method is initialized as k cluster centers and the clusters then iteratively refined. It converges when a saturation phase emerges where there is no further chance for changes in assignment of the clusters. The method aims to minimize a squared error function as an objective function defined in Eq.(15):

$$\underset{s}{\arg\min} \sum_{i=1}^{k} \sum_{\partial \in s_i} \|\partial - \mu_i\| \qquad (15)$$

where $\partial$ belongs to a set of observations ($\partial_1, \partial_2, \ldots, \partial_n$) in which each observation is a d-dimensional real vector, and $\mu$ is the cluster centers belongs to $S=\{S_1, S_2, \ldots, S_k\}$.

The utilized method, first, takes the original map, represented by an image of size 350-by-350 pixels where each pixel corresponds to $10 \times 10 \ m^2$, in which the allowed zones for the vehicle's operation are illustrated by the blue sections and the coastal areas are depicted with the brown color. Then, the clustering is applied and the operating and no-fly areas are transformed into the white and black regions in the clustered map, respectively. As a result, the planner works within the white (feasible) regions.

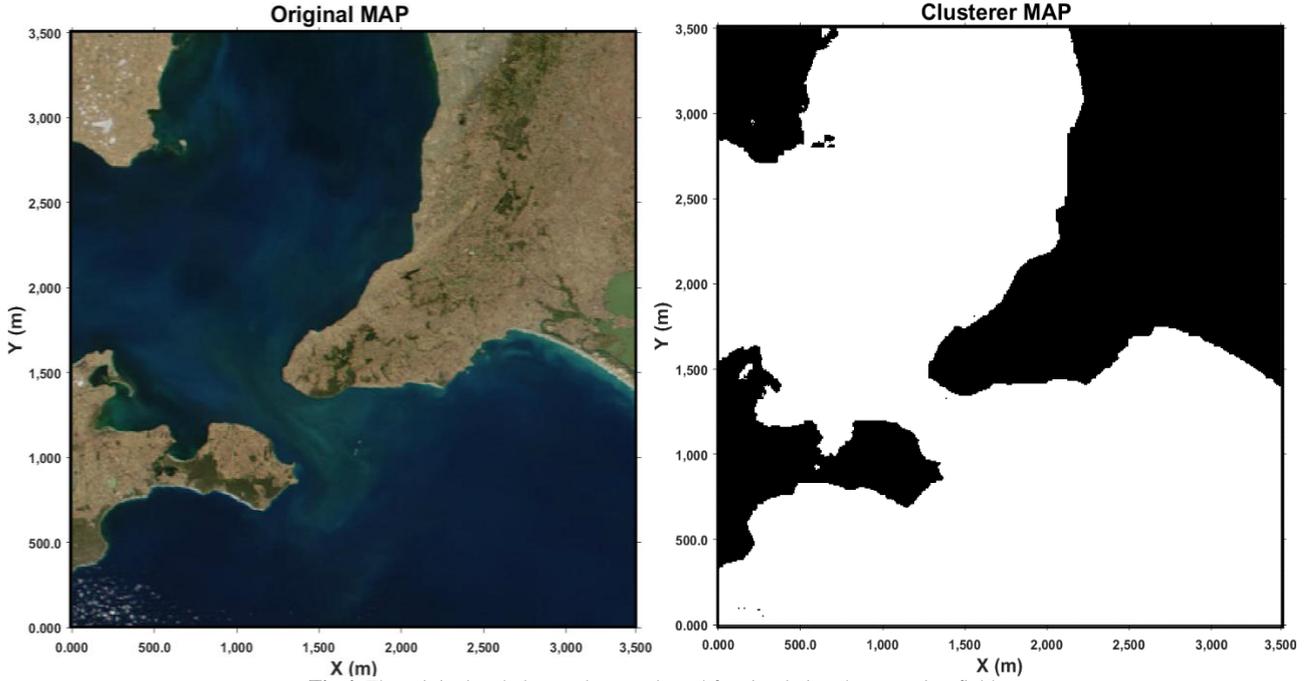

**Fig.4.** The original and clustered map selected for simulating the operating field

### 4.2 Mathematical Model with Uncertainty of Static/Dynamic Obstacles

Beside the offline map, where the coasts are assumed as static known obstacles, various uncertain objects are also considered in this study to encompass different possibilities of the real world situations. The AUV is equipped with sonar sensors for measuring the velocity and coordinates of the objects with a level of uncertainty depicted with a normal distribution. Modelling of different obstacles derived from [34-36] are formulated as follows:

**[1] Quasi-static Uncertain Objects**: Object's position should be placed between the location of AUV's starting and destination spot in the given map. This group of objects are introduced with a fixed-center generated at commencement and an uncertain radius that varying over time according to Eq.(16).

$$\forall \Theta_{x,y,z}, \quad \exists (\Theta_p, \Theta_r, \Theta_{Ur})$$
$$\Theta_p^i \in [(x_s, y_s, z_s), (x_d, y_d, z_d)] - \Theta_{r_{x,y,z}}^i$$
$$\Theta_p \sim \mathbf{N}(0, \sigma_1) \tag{16}$$
$$\Theta_{Ur} \sim \mathbf{U}(0, \sigma_2)$$
$$\Theta_r \sim \mathbf{N}(\Theta_p, \Theta_{Ur})$$

where $\Theta_p, \Theta_r, \Theta_{Ur}$ represent obstacles' position, radius and uncertainty, respectively. The $(x_s, y_s, z_s)$ and $(x_d, y_d, z_d)$ are the location of start and destination.

**[2] Moving Uncertain Objects:** This group of objects move spontaneously to random position in the operating environment, in which case the uncertainty encountered on objects position changes over the time according to Eq.(17).

$$\Theta_p(t) = \Theta_p(t-1) \pm \mathbf{U}(\Theta_{p_0}, \Theta_{Ur}) \tag{17}$$

**[3] Dynamic Uncertain Obstacles:** This group of objects moving self-propelled while their motion gets affected by current fields and calculated according to Eq.(18).

$$U_R^C = |V_C| \sim \mathbf{N}(0, 0.3)$$
$$X_{(t-1)} \sim \mathbf{N}(0, \sigma_3)$$
$$\Theta_p(t) = \Theta_p(t-1) \pm \mathbf{N}(\Theta_{p_0}, \Theta_{Ur}) \tag{18}$$
$$\Theta_r(t) = B_1 \Theta_r(t-1) + B_2 X_{(t-1)} + B_3 \Theta_{Ur}$$
$$B_1 = \begin{bmatrix} 1 & U_R^C(t) & 0 \\ 0 & 1 & 0 \\ 0 & 0 & 1 \end{bmatrix}, B_2 = \begin{bmatrix} 0 \\ 1 \\ 1 \end{bmatrix}, B_3 = \begin{bmatrix} 0 \\ 0 \\ U_R^C(t) \end{bmatrix}$$

where the $U_R^C$ is the impact of current on objects motion.

### 4.3 Mathematical Model of Static/Dynamic Current Field

This study uses a numerical estimator based on multiple Lamb vortices and Navier-Stokes equations [37] for modeling ocean current behavior in 3D space. Two types of static and dynamic current flow are mentioned in the following.

[1] The AUV's deployment usually is assumed on horizontal plane, because vertical motions in ocean structure are generally negligible due to large horizontal scales comparing vertical [37]. Hence, the physical model employed by the AUV to identify the current velocity field $V_C=(u_c,v_c)$ mathematically described by Eq.(19)-(20):

$$\frac{\partial \omega}{\partial t} + (\vec{V}_C \nabla)\omega = \nu \Delta \omega$$

$$\omega(\vec{S}) = \frac{\Im}{\pi \ell^2} e^{\frac{-(\vec{S}-\vec{S}^O)^2}{\ell^2}} \quad (19)$$

$$u_c(\vec{S}) = -\Im \frac{y-y_0}{2\pi(\vec{S}-\vec{S}^O)^2} \left[1 - e^{\frac{-(\vec{S}-\vec{S}^O)^2}{\ell^2}}\right]$$

$$v_c(\vec{S}) = \Im \frac{x-x_0}{2\pi(\vec{S}-\vec{S}^O)^2} \left[1 - e^{\frac{-(\vec{S}-\vec{S}^O)^2}{\ell^2}}\right] \quad (20)$$

where the $\nu$ represents the viscosity of the fluid, $\omega$ gives the vorticity, $\Delta$ and $\nabla$ are the Laplacian and gradient operators, respectively; $S$ is a 2D spatial space, $S^o$ is the center of the vortex, $\ell$ is the radius of the vortex, and $\Im$ is the strength of the vortex.

[2] A 3D turbulent dynamic current field, is estimated by a multiple layer structure based on the generated 2D current. The currents circulation patterns gradually change with depth. Therefore, to estimate continuous circulation patterns of each subsequent layer, a recursive application of Gaussian noise is applied to the parameters of 2D case. A probability density function of the multivariate normal distribution is employed to calculate the vertical profile $w_c$ of the 3D current $V_C=(u_c,v_c,w_c)$, which is given by Eq.(21):

$$w_c(\vec{S}) = \gamma \Im \frac{1}{\sqrt{\det(2\pi \hat{\lambda}_w)}} \times e^{\frac{-(\vec{S}-\vec{S}^o)^T}{2\hat{\lambda}_w(\vec{S}-\vec{S}^o)}}, \hat{\lambda}_w = \begin{bmatrix} \ell & 0 \\ 0 & \ell \end{bmatrix} \quad (21)$$

where $\lambda_w$ is a covariance matrix of the vortex radius $\ell$. A parameter $\gamma$ is used to scale the $w_c$ from the current horizontal profile $u_c$ and $v_c$ due to weak vertical motions of ocean environment. For generating dynamic time varying ocean current, Gaussian noise can be applied on $S^o, \Im,$ and $\ell$ parameters recursively as defined in Eq.(22):

$$V_c = f(\vec{S}^O, \Im, \varsigma)$$

$$S_i^o = A_1 S_{i-1}^o + A_2 X_{(i-1)}^{S_x} + A_3 X_{(i-1)}^{S_y}$$

$$\ell_{(i)} = A_1 \ell_{(i-1)} + A_2 X_{(i-1)}^\ell \quad (22)$$

$$\Im_{(i)} = A_1 \Im_{(i-1)} + A_2 X_{(i-1)}^\Im$$

$$A_1 = \begin{bmatrix} 1 & 0 \\ 0 & 1 \end{bmatrix}, A_2 = \begin{bmatrix} U_R^C(t) \\ 0 \end{bmatrix}, A_3 = \begin{bmatrix} 0 \\ U_R^C(t) \end{bmatrix}$$

where $U_R^C(t)$ is the update rate of current field in time $t$, and the rest of the unknown parameters in Eq.(22) are defined based on normal distribution as follows: $X^{S_x}_{(i-1)} \sim N(0,\sigma_{Sx})$, $X^{S_y}_{(i-1)} \sim N(0,\sigma_{Sy})$, $X^\ell_{(i-1)} \sim N(0,\sigma_\ell)$, $X^\Im_{(i-1)} \sim N(0,\sigma_\Im)$. The current field continuously updated every 4s during the AUV path planning process.

## 5 Review of Evolutionary Algorithms For Path Planning

Before going through the brief review of the employed evolutionary algorithms for the proposed rendezvous path planning, it is noteworthy to mention how they are applied over the optimization problem. As mentioned before, B-Spline curves are exploited to parameterize the path. The generated path by B-Spline curves captured from a set of control points like $\vartheta = \{\vartheta_1, \vartheta_2,...,\vartheta_i...,\vartheta_n\}$ in the problem space with coordinates of $\vartheta_1:(x_1,y_1,z_1),..., \vartheta_n:(x_n,y_n,z_n)$, where $n$ is the number of corresponding control points. Therefore, appropriate location of these control points play a substantial role in determining the optimal path. The mathematical description of the B-Spline coordinates is given by Eq.(23):

$$X(t) = \sum_{i=1}^{n} \vartheta_{x(i)} B_{i,K}(t)$$

$$Y(t) = \sum_{i=1}^{n} \vartheta_{y(i)} B_{i,K}(t) \quad (23)$$

$$Z(t) = \sum_{i=1}^{n} \vartheta_{z(i)} B_{i,K}(t)$$

Where $\vartheta_i$ is an arbitrary control point along the path ($\wp$), $B_{i,K}(t)$ is the curve's blending functions, $t$ is the time step, and $K$ is the smoothness factor. Further mathematical description of B-Spline curves can be found in [38].

In this regard, all control points should be located in a respective search region constraint to a predefined bounds of $\beta^i_\vartheta = [L^i_\vartheta, U^i_\vartheta]$ where $L^i_\vartheta$ and $U^i_\vartheta$ are the is the lower and upper bounds respectively, that for a control point like $\vartheta_i:(x_i,y_i,z_i)$ in the Cartesian coordinates are defined as Eq.(24):

$$L_{\vartheta(x)} = [x_0, x_1, x_2, ..., x_{i-1}, ..., x_{n-1}], \; U_{\vartheta(x)} = [x_1, x_2, ..., x_i, ..., x_n],$$
$$L_{\vartheta(y)} = [y_0, y_1, y_2, ..., y_{i-1}, ..., y_{n-1}], U_{\vartheta(y)} = [y_1, y_2, ..., y_i, ..., y_n], \quad (24)$$
$$L_{\vartheta(z)} = [z_0, z_1, z_2, ..., z_{i-1}, ..., z_{n-1}], \; U_{\vartheta(z)} = [z_1, z_2, ..., z_i, ..., z_n],$$

where $(x_0,y_0,z_0)$ and $(x_n,y_n,z_n)$ are the position of the *start* and *target* points, respectively. For the purpose of the initialization in the optimization process in this study, each control point $\vartheta_i:(x_i,y_i,z_i)$ is generated by Eq.(25):

$$x_i(t) = L^i_{\vartheta(x)} + Rand^x_i(U^i_{\vartheta(x)} - L^i_{\vartheta(x)})$$
$$y_i(t) = L^i_{\vartheta(y)} + Rand^y_i(U^i_{\vartheta(y)} - L^i_{\vartheta(y)}) \quad (25)$$
$$z_i(t) = L^i_{\vartheta(z)} + Rand^z_i(U^i_{\vartheta(z)} - L^i_{\vartheta(z)})$$

where *Rand* represents a random function. The ultimate goal of the optimization process is to find the best B-spline's control points in the problem space to have an optimal path satisfying the proposed augmented objective function Eq.(14).

Application of different metaheuristic methods may result very diverse on a same problem due to specific nature each problem; hence, accurate selection of the algorithm and proper matching of the algorithms functionalities according to nature of a particular problem is important issue to be considered, which usually remained unattended in most of the engineering and robotics frameworks in both theory and practice.

### 5.1 Particle Swarm Optimization

The PSO is an optimization method that performs fast computation and efficient performance in solving variety of the complex problems and widely used in past [39, 22, 34]. This algorithm is well suited on path planning problem due to its efficiency in handling complex and multi-objective problems. The PSO start operating with initial population of particles, where particles involve position and velocity in the search space. The position ($\chi_i$) and velocity ($\upsilon_i$) of each particle gets updated iteratively. Then, the performance of particles evaluated according to the fitness/cost functions. Each particle has memory to save previous state values and the best position in its experience as the. Then, the performance of particles evaluated according to the fitness/cost functions. Particles preserve their previous state value, personal ($\chi^{P-best}$) and global best ($\chi^{G-best}$) positions. Particle position and velocity gets updated using Eq.(26):

$$\upsilon_{ij}(t) = \omega \upsilon_{ij}(t-1) + c_1 r_1 \left[\chi^{P-best}_{ij}(t-1) - \chi_{ij}(t-1)\right] + c_2 r_2 \left[\chi^{G-best}_{ij}(t-1) - \chi_{ij}(t-1)\right] \quad (26)$$
$$\chi_{ij}(t) = \chi_{ij}(t-1) + \upsilon_{ij}(t)$$

where $c_1, c_2$ are acceleration parameters; $r_1, r_2$ are randomness factors; $\omega$ is weight for balancing the local-global search. More explanation of PSO process can be found in [39].

Particles are coded by coordinates of a potential path generated by B-Spline curves. The position and velocity parameters of the particles correspond to the coordinates of the B-Spline control points ($\vartheta$) utilized in path generation. As the PSO iterates, each particle is attracted towards its respective local attractor based on the outcome of the particle's individual and swarm search. The pseudo code of the PSO algorithm and its mechanism on path planning process is provided in Fig.5. The most attractive benefit of the PSO is that it is easy to implement and it requires less computation to converge to the optimum solution. However, main drawback of PSO is that the particles rapidly get converged so that it easily traps in a local optima.

```
PSO Path Planning
Initialize each particle by random velocity and position in following steps:
  ▪ Assign B-Spline control points ϑ_i as particle position χ_i
  ▪ Initialize each particle with random velocity υ_i in range of predefined bounds β^i_ϑ=[U^i_ϑ,L^i_ϑ]
  ▪ Choose appropriate parameters for the population size (i_max)
  ▪ Set the number of control-points m used to generate the B-Spline path
  ▪ Set the maximum number of iterations (t_max)
  ▪ Initialize χ_i^{P-best}(1) with current position of each particle at first iteration t=1.
  ▪ Set the χ^{G-best}(1) with the best particle in initial population at t=1.
For  t=1 to t_max
  Evaluate each candidate particle according to given cost function
  For  i=1 to i_max
    Updated the particles χ_i^{P-best} and χ^{G-best} at iteration t
    if  Cost_φ(χ_i(t)) ≤ Cost_φ(χ_i^{P-best}(t-1))
       χ_i^{P-best}(t)= χ_i(t)
    else
       χ_i^{P-best}(t)= χ_i^{P-best}(t-1)
    end (if)
    χ^{G-best}(t) = arg min_{1≤i} Cost_φ(χ_i^{P-best}(t))
    Update the state of the particle in the swarm
    υ_i(t) = ωυ_i(t-1) + c_1r_1[χ_i^{P-best}(t-1) - χ_i(t-1)] + c_2r_2[χ^{G-best}(t-1) - χ_i(t-1)]
    χ_i(t) = χ_i(t-1) + υ_i(t)
    Evaluate each candidate particle χ_i according to given cost function Cost_φ(χ_i(t))
  end (For)
  Transfer best particles to next generation
end (For)
Output χ^{G-best}  and its correlated path as the optimal solution
```

**Fig.5.** PSO optimal path planning

### 5.2 Biogeography-based Optimization

The Biogeography-based Optimization (BBO) is an evolution based technique that mimics the equilibrium pattern of inhabitancy in biogeographical island [40]. This algorithm follows the species' immigration and emigration phenomena. The BBO starts its process with initial population of candidate solutions known as habitats. This algorithm has strong exploitation ability as the habitats never get eliminated from the population but get improved by migration. The fitness of the habitats in BBO is presented using a performance factor of habitat suitability index (HSI). High HSI solutions correspond to habitats with more suitable habitation that tend to share their useful information with poor HIS solutions. There is also a qualitative parameter of Suitability Index Variable (SIV) that is a vector of integers and affects habitability of solutions. The SIV initialized in advance with random integers. Each candidate solution of habitat ($h_i$) holds migration parameters of emigration rate (µ), immigration rate (λ), SIV vector; and gets evaluated by its corresponding HIS value before the BBO start its process. A good solution has a higher rate of µ (emigration) and lower rate of λ (immigration). The SIV vector of $h_i$ is probabilistically altered according to its immigration rate of λ. In migration process, one of the solutions ($h_i$) is randomly picked to migrate its SIV to another solution $h_j$ regarding its µ value. Then, mutation is exerted to alter the population and prevent stuck in a local optima. A candidate solution of $h_i$ is changed and modified considering probability of habitance of $S$ species in $h_i$ at time $t$; so its habitance probability at time ($t+\Delta t$) is determined using Eq.(28) by holding one of the conditions given by Eq.(27) as follows:

$$\forall h_i(t) \quad \exists \lambda_S, \mu_S, P_S(t)$$
$$\begin{cases} \forall h_i(t):\exists S & \Rightarrow & \forall h_i(t+\Delta t):\exists S \\ \forall h_i(t):\exists S & \Rightarrow & \forall h_i(t+\Delta t):\exists S-1 \\ \forall h_i(t):\exists S & \Rightarrow & \forall h_i(t+\Delta t):\exists S+1 \end{cases} \quad (27)$$

$$P_S(t+\Delta t) = P_S(t)(1-\lambda_S \Delta t - \mu_S \Delta t) + P_{S-1}\lambda_{S-1}\Delta t + P_{S+1}\mu_{S+1}\Delta t \quad (28)$$

$$\lambda_S = I*\left(1-\frac{S}{S_{\max}}\right)$$
$$\mu_S = E*\left(\frac{S}{S_{\max}}\right) \quad ;\xrightarrow{if\ E=I} \lambda_S+\mu_S = E$$

where $E$ and $I$ are the maximum emigration and immigration rates, respectively. $S_{max}$ is the maximum number of species in a habitat. Assuming a very small $\Delta t \approx 0$, the $P_s$ is calculated by Eq.(29).

$$\dot{P_S} = \begin{cases} -(\lambda_S+\mu_S)P_S + P_{S+1}\lambda_{S+1} & S=0 \\ -(\lambda_S+\mu_S)P_S + P_{S+1}\mu_{S+1} + P_{S-1}\lambda_{S-1} & 1 \le S \le S_{\max}-1 \\ -(\lambda_S+\mu_S)P_S + P_{S-1}\mu_{S-1} & S=S_{\max} \end{cases} \quad (29)$$

Solutions with smaller $P_s$ need to be mutated, while habitats with higher $P_s$ is less likely to mutate; thus, in BBO process, the mutation rate of $m(S)$ has inverse relation to habitats $P_s$ value as indicated by Eq.(30).

$$m(S) = m_{\max}\left[\frac{1-P_S}{P_{\max}}\right] \quad (30)$$

In Eq.(30), the $P_{max}$ is probability of the habitat with $S_{max}$ and $m_{max}$ represents the maximum mutation rate. Further details can be found in [35].

In the proposed BBO based path planner, each habitat $h_i$ corresponds to the coordinates of the B-Spline control point's $\vartheta_i$ utilized in path generation, where $h_i$ defined as a parameter to be optimized ($P_s:(h_1,h_2,…,h_{n-1})$). Habitats get improved iteratively. The pseudo code of the BBO on path planning process is provided in Fig.6.

### 5.3 Firefly Algorithm

Firefly Algorithm (FA) is a kind of meta-heuristic algorithm [41] captured by sampling the flashing patterns of fireflies. In the FA process, the fireflies are attracted to each other according to their brightness, which it gets faded when fireflies' distance is increased. The brighter fireflies' attract the others in the population. The firefly's attraction depends on their distance $L$ and received brightness from the adjacent fireflies; thus, the firefly $\chi_i$ approaches the $\chi_j$ according to their attraction parameter $\beta$ that is calculated by Eq.(31):

$$L_{ij} = \|\chi_j - \chi_i\| = \sqrt{\sum_{q=1}^{d}(x_{i,q}-x_{j,q})^2}$$
$$\beta = \beta_0 e^{-\gamma L^2} \quad (31)$$
$$\chi_i^{t+1} = \chi_i^t + \beta_0 e^{-\gamma L_{ij}^2}(\chi_j^t - \chi_i^t) + [rand - \frac{1}{2}]$$
$$\alpha_t = \alpha_0 \delta^t, \delta \in (0,1)$$

---

**BBO based Path Planning**
Initialize a set of solutions as initial habitat population
- Assign B-spline control points $\vartheta_i$ as habitat $h_i$
- Choose appropriate parameters for the population size $i_{max}$
- Set the number of control-points ($n$) of the B-Spline
- Set the maximum number of iteration $t_{max}$
- Assign maximum immigration and emigration rate ($I$, $E$)
- Assign maximum mutation rate $m_{max}$
- Set $S_{max}$ and SIV vector

**For** $t=1$ **to** $t_{max}$
  Compute λ and µ rates for each solution
  Compute hesitance probability for each solution
  Evaluate the fitness (HSI) of each habitat
  Modify habitats based on λ and µ (Migration)
  **For** $i=1$ **to** $i_{max}$
    Use $\lambda_i$ to probabilistically decide whether immigrate to $h_i$
    **if** $rand(0,1) < \lambda_i$
      **For** $j=1$ **to** $i_{max}$
        Select the emigrating habitat $h_j$ with probability $\propto \mu_j$
        **if** $rand(0,1) < \mu_j$
          Replace a randomly selected SIV variable of $h_i$ with its value in $h_j$
        *end*
      *end*
    *end*
  *end*
  Carry out the mutation
  Transfer the best solution to the next generation
*end*
Output the best habitat and its correlated path as the optimal solution

**Fig.6.** BBO optimal path planning

---

**FA based Path Planner**
Initialization phase:
- Initialize population of fireflies $\chi_i (i=1,2,…,n)$ with the control points $\vartheta_i$
- Set the number of population $i_{max}$
- Define light absorption coefficient γ
- Initialize the attraction coefficient $\beta_0$
- Set the damping factor of δ
- Initialize the randomness scaling factor of $\alpha_0$
- Set the parameter of randomization $\alpha_t$
- Set the maximum iteration $t_{max}$

**For** $t=1: t_{max}$
  **For** $i=1: i_{max}$
    Reconstruct a path according to $\chi_i$
    Evaluate the path $Cost_{fp}(\chi_i(t))$
    Update light intensity of $\chi_i$
    **For** $j=1:i$
      Reconstruct a path according to $\chi_j$
      Evaluate the path $Cost_{fp}(\chi_j(t))$
      Update light intensity of $\chi_j$
      **If** ($\beta j > \beta i$),
        Move firefly $i$ towards $j$
      *end*
    *end*
  *end*
  Rank the fireflies and find the current best
*end*
Output result

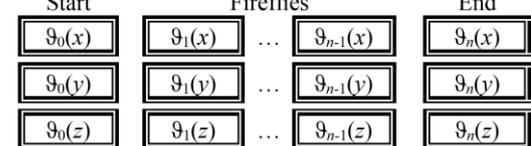

Fig.7. Pseudo-code of FA mechanism on path planning approach and encoding scheme based on the path control points

In Eq.(31), the coordinate firefly $\chi_i$ in d dimension space contains $q$ components of $x_{i,q}$. The $\beta_0$ is attraction value of firefly $\chi_i$ when $L=0$. The $\alpha_t$ and $\alpha_0$ are the randomization factor and initial randomness scaling value, respectively. The $\gamma$ and $\delta$ are light absorption coefficient and damping factor, respectively. It is noteworthy to mention that there should be a suitable balance between mentioned parameters because in a case that $\gamma \rightarrow 0$ the algorithm mechanism turns to particle swarm optimization, while $\beta_0 \rightarrow 0$ turns the movement of fireflies to a simple random walk [41]. In path planning procedure, the fireflies correspond to B-Spline control points ($\{\chi_1=\vartheta_1^{x,y,z}, \ldots, \chi_i=\vartheta_i^{x,y,z}, \ldots, \chi_n=\vartheta_n^{x,y,z}\}$) tends to be optimized toward the best solution in the search space. The FA is more efficient comparing others as it is privileged to apply an automatic subdivision approach that improves its capabilities in dealing with highly nonlinear and multi-objective problems [42]. The attraction has inverse relation with distance; therefore, the firefly population gets subdivided and subgroups swarm locally in parallel toward the global optima, which this mechanism enhances convergence rate of the algorithm. The pseudo-code of FA based path planning process is provided by Fig.7.

### 5.4 Differential Evolution

Differential Evolution (DE) algorithm is a population-based optimization algorithm that introduced as a modified version of genetic algorithm and applies similar evolution operators (crossover, mutation and selection) [43]. DE uses floating numbers and real coding for presenting problem parameters that enhances solution quality and provides faster optimization. The algorithm uses a non-uniform crossover and differential mutation; the uses selection operator to propel the solutions toward the optimum area in the search space. The mechanism of the DE algorithm is given below:

[1] **Initialization:** In this step the population initialized with solution vectors $\chi_i$, ($i=1, \ldots, i_{max}$), in which any vector $\chi_i^{x,y,z}$ is assigned with an arbitrary path $\wp^i_{x,y,z}$ while the control points $\vartheta$ correspond to elements of $\chi^{x,y,z}$ vector as given by Eq.(32):

$$\forall i \in \{1,\ldots,i_{max}\}$$
$$\forall t \in \{1,\ldots,t_{max}\}$$
$$\chi_{i,t}(\chi_{i,t}^x, \chi_{i,t}^y, \chi_{i,t}^z) \equiv (\vartheta_x^i(t), \vartheta_y^i(t), \vartheta_z^i(t)) \quad (32)$$

where $i_{max}$ is the population size, and $t_{max}$ is the maximum number of iterations.

[2] **Mutation:** The main reason for superior performance of DE comparing to GA is using an efficacious mutation scheme. Three solution vector like $\chi_{r1,t}$, $\chi_{r2,t}$, and $\chi_{r3,t}$ are picked randomly in iteration $t$ and the difference vector between them utilized in mutation procedure as given by Eq.(33). This kind of mutation uses a *donor* parameter, which *donor* is one of the randomly selected triplet and accelerates rate of convergence, given by Eq.(34):

$$\dot{\chi}_{i,t} = \chi_{r3,t} + S_f(\chi_{r1,t} - \chi_{r2,t})$$
$$r1, r2, r3 \in \{1,\ldots,i_{max}\} \quad (33)$$
$$r1 \neq r2 \neq r3 \neq i, \; F \in [0,1+]$$

$$donor = \sum_{i=1}^{3} \left( \lambda_i \bigg/ \sum_{j=1}^{3} \lambda_j \right) \chi_{ri,G} \quad (34)$$

where $\dot{\chi}_{i,t}$ and $\chi_{i,t}$ are mutant and parent solution vector, respectively. $S_f$ is a scaling factor that balances the difference vector ($\chi_{r1,t} - \chi_{r2,t}$) and higher $S_f$ enhances algorithm's exploration capability. $\lambda_j \in [0,1]$ is a uniformly distributed parameter.

[3] **Crossover:** This operator shuffles the mutant individuals and the existing population members using Eq.(35):

$$\begin{cases} \chi_{i,t} = (x_{1,i,t}, \ldots, x_{n,i,t}) \\ \dot{\chi}_{i,t} = (\dot{x}_{1,i,t}, \ldots, \dot{x}_{n,i,t}) \\ \ddot{\chi}_{i,t} = (\ddot{x}_{1,i,t}, \ldots, \ddot{x}_{n,i,t}) \end{cases}$$
$$\ddot{x}_{j,i,t} = \begin{cases} \dot{x}_{j,i,t} & rand_j \leq r_C \vee j=k \\ x_{j,i,t} & rand_j \leq r_C \wedge j \neq k \end{cases} \quad (35)$$
$$j=1,\ldots,n; \quad n \in [1, i_{max}]$$

where $\ddot{\chi}_{i,t}$ is the produced offspring and $r_C \in [0,1]$ is the crossover rate.

[4] **Validation and Selection:** The solutions produced by the mutation and crossover get evaluated and the best solutions are shifted to the next generation ($t+1$). All solutions are evaluated by predefined cost function and the worst solutions get discarded from the population. The selection procedure is given by Eq.(36) and the

```
DE based Path Planning
Initialization phase:
  ▪ Initialize population of solution vectors randomly χᵢˣ'ʸ'ᶻ with the control
    points (ϑⁱₓ,ϑⁱᵧ,ϑⁱᵤ)
  ▪ Set the maximum number of iteration tₘₐₓ
  ▪ Choose appropriate parameters for the population size iₘₐₓ
  ▪ Set the mutation coefficient r_C
  For t =1 to tₘₐₓ
    For i =1 to iₘₐₓ
      Reconstruct a path according to χᵢ,ₜ
      Evaluate the path
      Determine the donor
      Apply mutation using
         χ̇ᵢ,ₜ = χᵣ₃,ₜ + S_f(χᵣ₁,ₜ − χᵣ₂,ₜ)
         r1,r2,r3 ∈ {1,...,iₘₐₓ}, r1 ≠ r2 ≠ i, S_f ∈ [0,1+]
      For j =1 to i
        Apply crossover using χᵢ,ₜ mutant solution χ̇ᵢ,ₜ to get the χ̈ᵢ,ₜ
        Reconstruct a path according to χᵢ,ₜ, χ̇ᵢ,ₜ and χ̈ᵢ,ₜ
        Evaluate the corresponding paths to χᵢ,ₜ, χ̇ᵢ,ₜ and χ̈ᵢ,ₜ
        if Cost_℘(χᵢ,ₜ) ≤ Cost_℘(χ̇ᵢ,ₜ)
          χ̇ᵢ,ₜ₊₁ = χᵢ,ₜ
          if Cost_℘(χᵢ,ₜ) ≤ Cost_℘(χ̈ᵢ,ₜ)
            χ̇ᵢ,ₜ₊₁ = χᵢ,ₜ
          else
            χ̇ᵢ,ₜ₊₁ = χ̈ᵢ,ₜ
          end
        else
          χ̇ᵢ,ₜ₊₁ = χ̇ᵢ,ₜ
        end
      end
    end
    Select the best solutions to transfer to next iteration
  end
  Output best solution and the corresponding paths
```

Fig.8. Pseudo code of DE based path planning

process of DE to the proposed path planning problem is expressed by Fig.8.

$$\dot{\chi}_{i,t+1} = \begin{cases} \dot{\chi}_{i,t} & Cost_\wp(\dot{\chi}_{i,t}) \leq Cost_\wp(\chi_{i,t}) \\ \chi_{i,t} & Cost_\wp(\dot{\chi}_{i,t}) > Cost_\wp(\chi_{i,t}) \end{cases}$$

$$\ddot{\chi}_{i,t+1} = \begin{cases} \ddot{\chi}_{i,t} & Cost_\wp(\ddot{\chi}_{i,t}) \leq Cost_\wp(\chi_{i,t}) \\ \chi_{i,t} & Cost_\wp(\ddot{\chi}_{i,t}) > Cost_\wp(\chi_{i,t}) \end{cases} \quad (36)$$

## 6  Simulation Results and Discussion

There is a significant distinction between theoretical understanding of evolutionary algorithms and their properness for applying on different problems due to the size, complexity, and nature of different problems. This may lead to different results using different algorithms, for a same problem. Therefore, careful selection of evolutionary method is of significant importance. In this part, the simulation results obtained for the rendezvous path planning problem through different scenarios are shown and analyzed. The objective is to show the performance of the utilized evolutionary methods for applying to the online path planning mechanism resulting in an adaptive maneuverability of the vehicle in a fully cluttered dynamic operating field. In each scenario, important aspects of uncertain and cluttered operating field are considered and the outcome of the planner module is demonstrated. In the last scenario, all possible challenges of a realistic operating field are implemented and full details of the planner outputs including the generated path, corresponding states' history and final rendezvous time, are presented. In the following, first, we describe the configuration of the utilized methods for the proposed problem and then we introduce the adopted scenarios to consider the performance of the evolutionary methods for online path planning. The parameter configurations of the algorithms are set as follows:

- **PSO:** The PSO is initialized by 100 particles (candidate paths). The expansion-contraction coefficients also set on *2.0* to *2.5*. The inertia weight varies from 1.4 to 0.5.
- **BBO:** The habitats population ($i_{max}$) is set on 100. The number of kept habitats is set on 40, number of new habitats is set on 40. The emigration rate is generated by a vector of μ including $i_{max}$ elements in (0,1), and the immigration rate is defined as λ=1- μ. The maximum mutation rate is set on 0.1.
- **FA:** Fireflies population set on 100, the attraction coefficient base value $β_0$ is set on 2, light absorption coefficient γ is assigned with 1. The damping factor of δ is assigned with 0.95 to 0.97. The scaling variations is defined based on initial randomness scaling factor of $α_0$. The parameter of randomization is set on 0.4.
- **DE:** Population size is set on 100, lower and upper bound of scaling factor is set on 0.2 and 0.8, respectively. The crossover probability fixed on 20 percent.

Meanwhile, the maximum number of iteration for all algorithms is set on $t_{max}$=100 and for having stable results the number of 30 runs for each scenario is applied. Finally, number of control points for each B-Spline path is set on 7.

**Scenario 1: Path planning in spatiotemporal ocean flows**

To investigate the performance of the proposed online path planning strategy, the ocean environment is modelled as a three dimensional volume $Γ_{3D}$ covered by uncertain static /moving obstacles and time varying ocean current, as mentioned in section 4. The update of operating field includes changes in the position of the obstacles or current behaviour, continuously measured from the on-board sonar and horizontal acoustic Doppler current profiler (HADCP) sensors. Fig.9 represents the performance of the path planners in adapting ocean currents that varies within 4 time steps (various updating rate) once the AUV starts to travel from a waypoint on a loitering pattern (shown with a red circle) toward the rendezvous position (indicated by a yellow square). In this scenario, the variability of the current field is illustrated through four subplots in Fig.9 labelled by Time Step 1-4. The current field used in Fig.9 is computed from a random distribution of 50 Lamb vortices in a 350×350 grid within a 2D spatial domain according to the pixel size of the clustered map. In order to change the current parameters, Gaussian noise, in a range of 0.1~0.8, is randomly applied to update current parameters of $S_i^o$, $ℓ$, and $ℑ$ given in Eq.(19)-(22).

Referring to Fig.9, it is apparent that the generated path by the evolutionary algorithms between the starting point (red circle) and the final rendezvous position (yellow square) adapts the behaviour of current in all time steps. As can be seen in each mentioned subplot, by changing the current map, the new information of spatiotemporal current flow, collected by HADCP, is fed to the online path planner module, and having a certain level of autonomy and adaptation, it generates a refined path that can avoid severely adverse current flows and can take advantage of favourable ones to speed up the vehicle motion and nose dive the energy expenditure. Being more specific in comparing the generated paths, obviously the FA method shows better performance in making use of favourable current flows, comparing to the others. In the second rank, the path generated by BBO shows significant flexibility in coping with current change specifically when the current magnitude gets sharper (more clear in turbulent given in Time Step: 3). Fig.10 demonstrates the variation of fitness value and the rate of solution convergence for the all evolutionary methods through this scenario. As can be seen, the convergence rate of the PSO is to somehow moderate and the rest of three methods are roughly identical in finding the optimal solution.

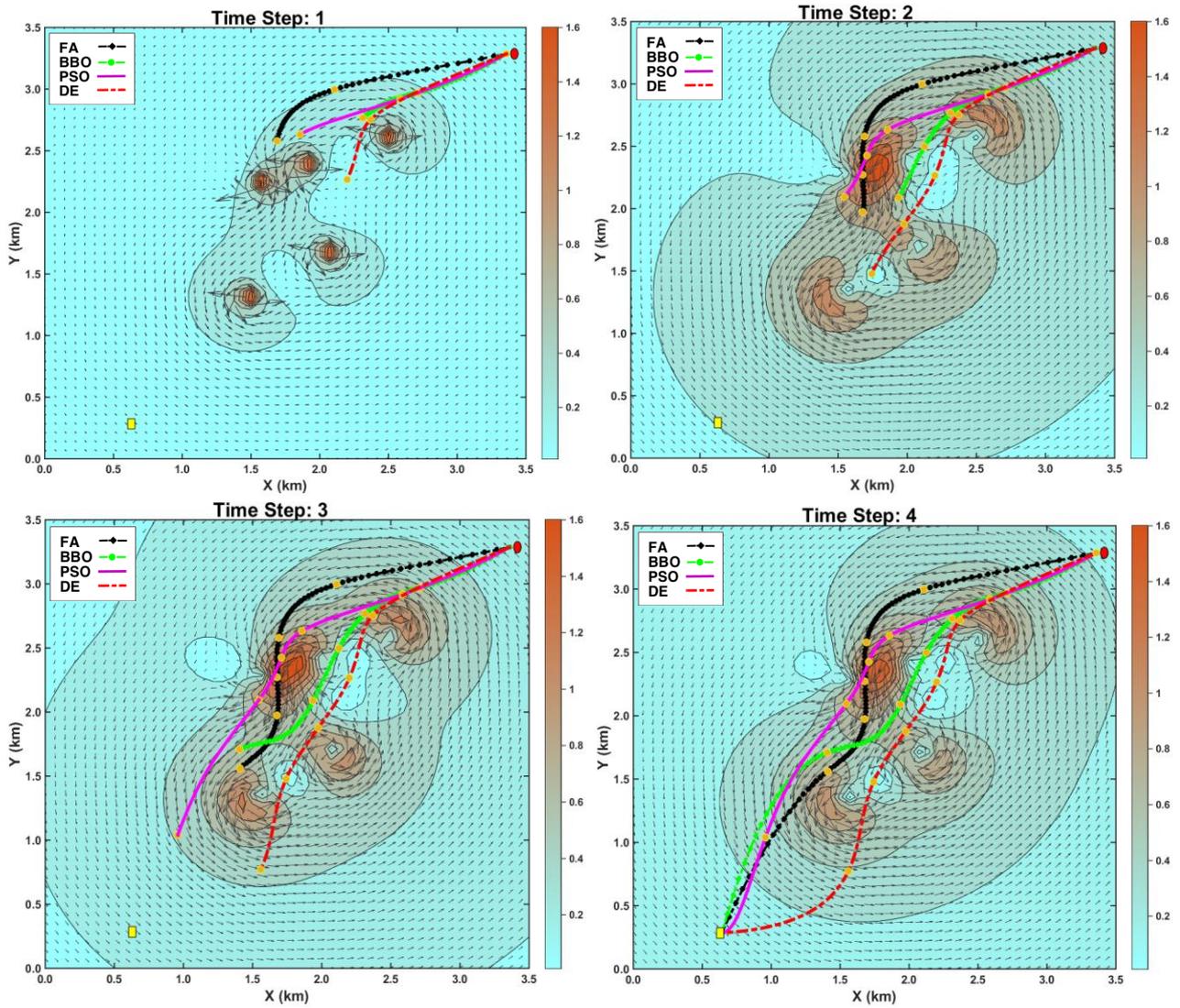

**Fig.9.** Adaptive path behaviour to current updates in 4 time steps

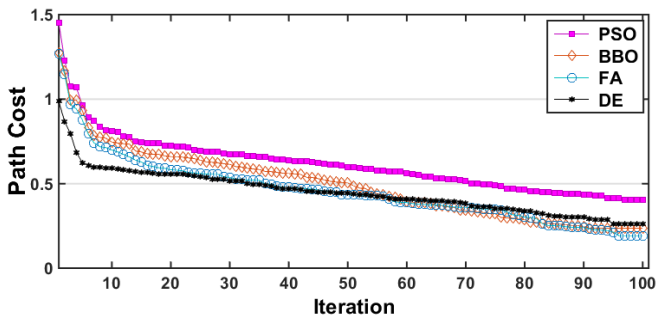

**Fig.10.** Variation of path cost and violation along the rendezvous mission

It is noteworthy to mention that setting lower rate for Gaussian noise parameter (i.e., in range of 0.1 to 0.8) applied to the current vector field parameters, leads better fitness values for the generated paths by the all four algorithms. Moreover, by setting higher current update rate (where Gaussian noise parameters get greater value than 0.8), the path fitness value reduces and practically other conditions remaining constant in path execution process. Thus, in a dynamic path planner, when the current field updating rate get significantly greater value than the path updating rate, the probability of path failure increases remarkably.

**Scenario 2: Path planning in a cluttered field with variable current flow and static obstacles**

In this scenario, the operating field is cluttered with the static obstacles, randomly distributed in the problem space. The variability of the current vector field is the same as the scenario 1. As depicted in Fig.11, distribution of the obstacles is in such a way that they occupy the solution space, used in scenario 1, for generating the path. In Fig.11(a), the planner module generates the candidate path considering the underlying current map and obstacles' situations toward the final rendezvous position. The current map is updated in the location where the 2D position of the vehicle is shown by the yellow triangles; these changes are obvious in Fig11(b). The thin curves in Fig.11(b) indicate the offline initial paths generated by the proposed methods (before updating the current map), and the refined paths (online paths), are clearly shown by the thick curves in Fig.11(c). The utilized FA, BBO, PSO and DE path planning methods are capable of generating a collision free path against the distribution of obstacles. Here again, the re-planning procedure is helpful to refine the offline path that may have not been tailored for the vehicle operation in the variable ocean flows. Fig.12 indicates the performance of the proposed methods in finding the optimal solution considering the augmented objective function in Eq.(14).

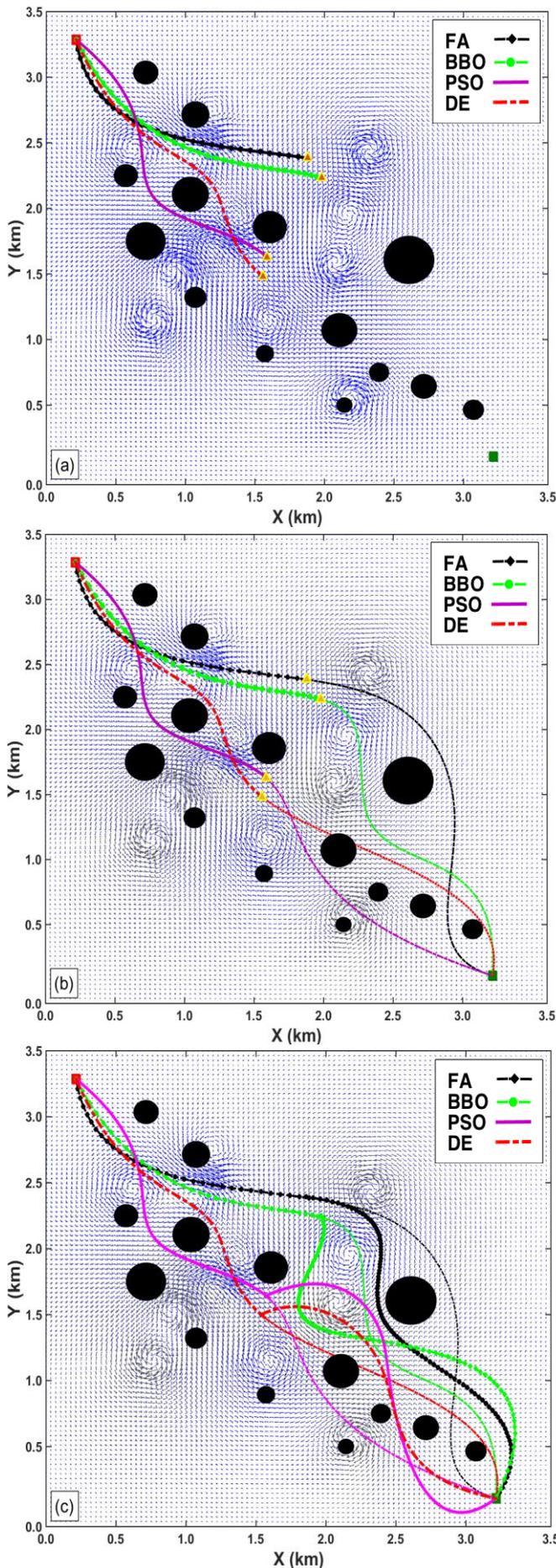

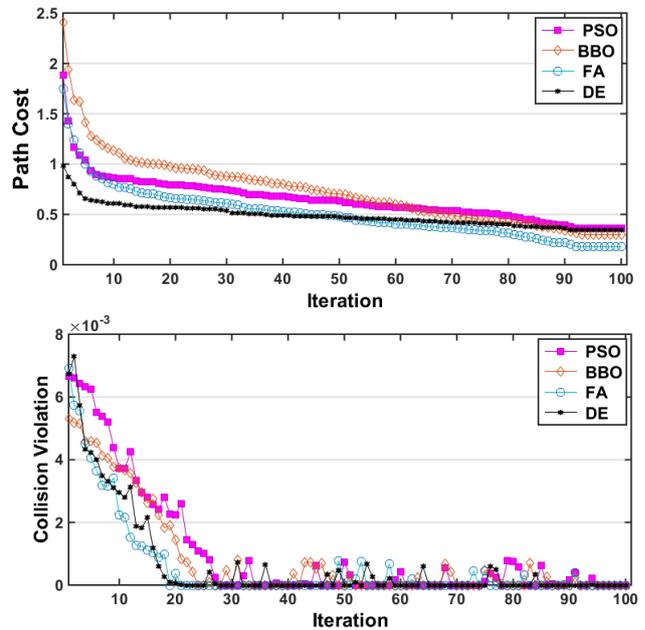

As can be seen in Fig.12, the initial solutions generated by the evolutionary methods are different, as a result of different natures and mechanisms. However, they all are capable of searching the optimization space and converge to the best possible minima. Refereeing to collision violation plot, it can be inferred that applying the re-planning strategy around 20th iteration, the violation value decreases to a great extent and at the end there is no violation over the obtained solution.

**Fig.12.** Variation of path cost and collision violation along the rendezvous mission

### Scenario 3: Path planning in a cluttered field with variable current flow static uncertain /moving obstacles

This scenario becomes more challenging as we add the uncertain static and moving obstacles to the variable spatiotemporal current field. The collision boundaries represented as green spheres around the obstacles computed with radius $\Theta_r \sim 2\sigma_o$ indicating a confidence of 98% that the obstacle is located within this area , referring to the equations mentioned in section 4. The proposed obstacles in this scenario are generated with random position by the mean and variance of their uncertainty distribution and velocity proportional to current velocity. The uncertainty over the both static and moving obstacles are linearly propagated relative to the updating time and leads to the radius growth of the static obstacles and simultaneous position and radius changes in the moving obstacles. As mentioned before, by using on-board sonar sensors the aforementioned variations of the obstacles are measured and tracked. Fig.13 illustrates 3D-path generated by the evolutionary methods for the scenario at hand. In this figure, for having a clear visualization, just the final refined path is indicated. As can be seen in this figure, due to the linear propagation of the uncertainty with time, there exist a collision boundary encircling the objects. The safe trajectory is achieved if the vehicle manoeuvre does not have any intersection with the proposed obstacle boundary. Obviously enough, the utilized evolutionary path planners are able to generate a collision free path through the dynamic obstacles that can satisfy the conditions of the proposed rendezvous problem. This fact is indicated by

**Fig.11.** Offline path behaviour and re-planning process in accordance with current and static obstacles

Fig.14 showing the expedient convergence rate of the methods in finding optimal solution and sharp convergence rate of collision violation, as iteration goes on.

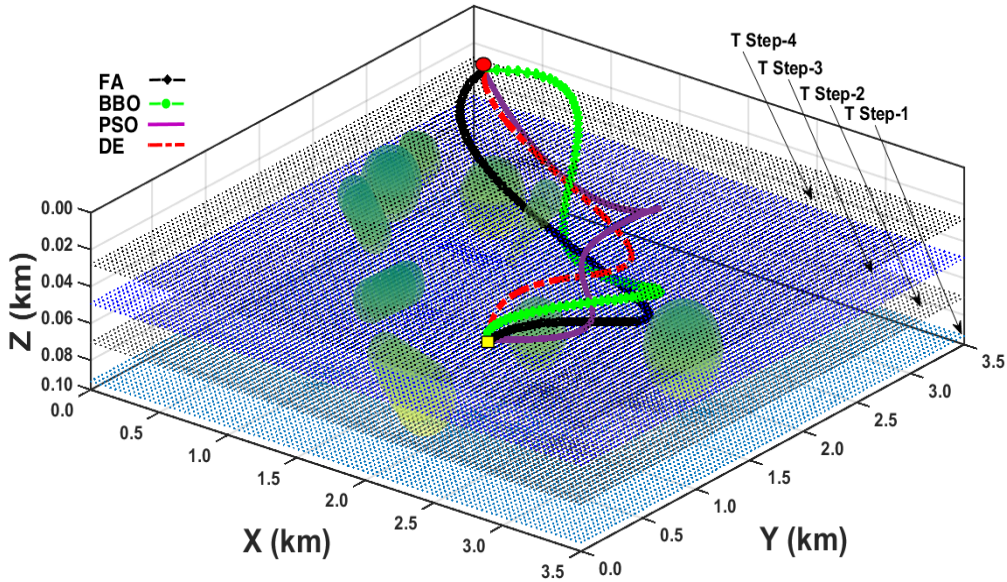

**Fig.13.** Evolution curves of FA, DE, BBO, and PSO algorithms with four time step change in water current flow in three-dimension space

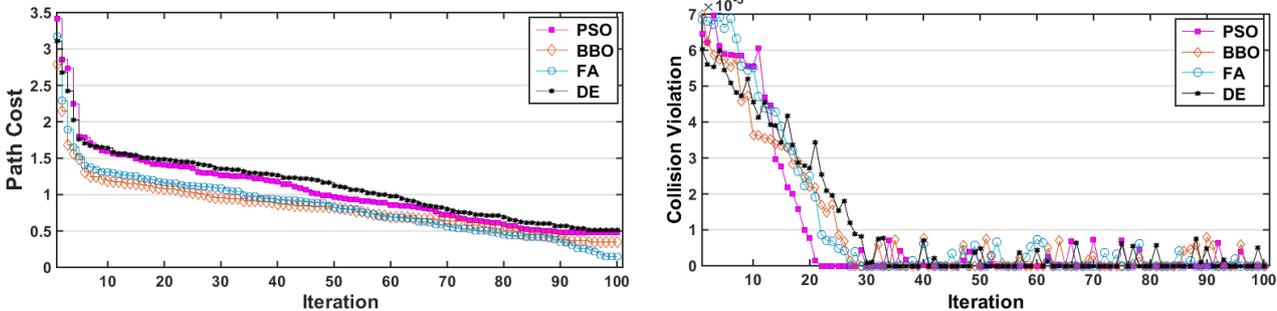

**Fig.14.** Variation of path cost and collision violation along the rendezvous mission

**Scenario 4: Path planning in a highly uncertain realistic operating field**

In the last scenario, the goal is to thoroughly simulate a highly uncertain cluttered realistic operating field including all possible barriers to assess the performance of the online path planner considering the rendezvous problem. In this scenario, the clustered map is used as an underlying environment embodied with variable ocean flows, uncertain static and moving obstacles. The coastal areas on the map are treated as no-fly zones and the planner should be aware of them. These zones in a context of an offline obstacles are fed to the planner module before starting the rendezvous mission. The leader AUV transmits its position, indicated by a yellow square in Fig.15, and the possible rendezvous time that is $T_r$=1800 (sec) in this scenario, to the follower counterpart. The threshold of $\varepsilon$=300 (sec) is permitted for the final rendezvous, as mentioned in (13). The follower AUV water-reference velocity is set on $\upsilon$=2.5 (m/s) and by which it starts the rendezvous mission from the red circle indicated on Fig.15. For the current flow, the vortexes radius ($\ell$) and strength parameters ($\Im$) have been set on as 2.8 m and 12 m/s, respectively [27]; the uncertain static and moving obstacles follows the same concept used in the previous scenario. Fig.15 (Time Step: 1) shows the firs path generated by the evolutionary methods. The yellow triangles indicate the place where the current vector field is updated and the planner uses the new situational awareness of the environment to refine the previous path. It is noteworthy to mention that the path generated by DE, indicated by red dash line, does not have collision with the neighbour coast at the start of the mission although in 2D illustration this may seem. The uncertainty of the obstacles can be clearly seen in the subsequent Time Step: 2-4 of Fig.15. It is obvious form Fig.15 that by severing the complexity of the problem, all the evolutionary path planners still generate safe collision free paths satisfying the conditions of the proposed NOCP. For example, Fig.16 demonstrates the final mission time achieved by the evolutionary path planners used in this study. As can be seen, the final time is different stemming from the inherent differences in structure and mechanism of the proposed methods; however, all of them properly satisfy the assigned rendezvous time threshold ($\varepsilon$=300 (sec)). The associated variations of the performance index and collision violation per iteration are depicted in Fig.17. Again, the convergence rate of collision violation to zero is sharp due to the nature of online planning strategy. Finally, Fig.18 illustrates the history of the states with respect to the vehicular constraints. As can obviously be inferred, the generated path by the all methods respect the physical constraints over the states and generates smooth trajectory that is applicable for the vehicle low-level auto pilot module.

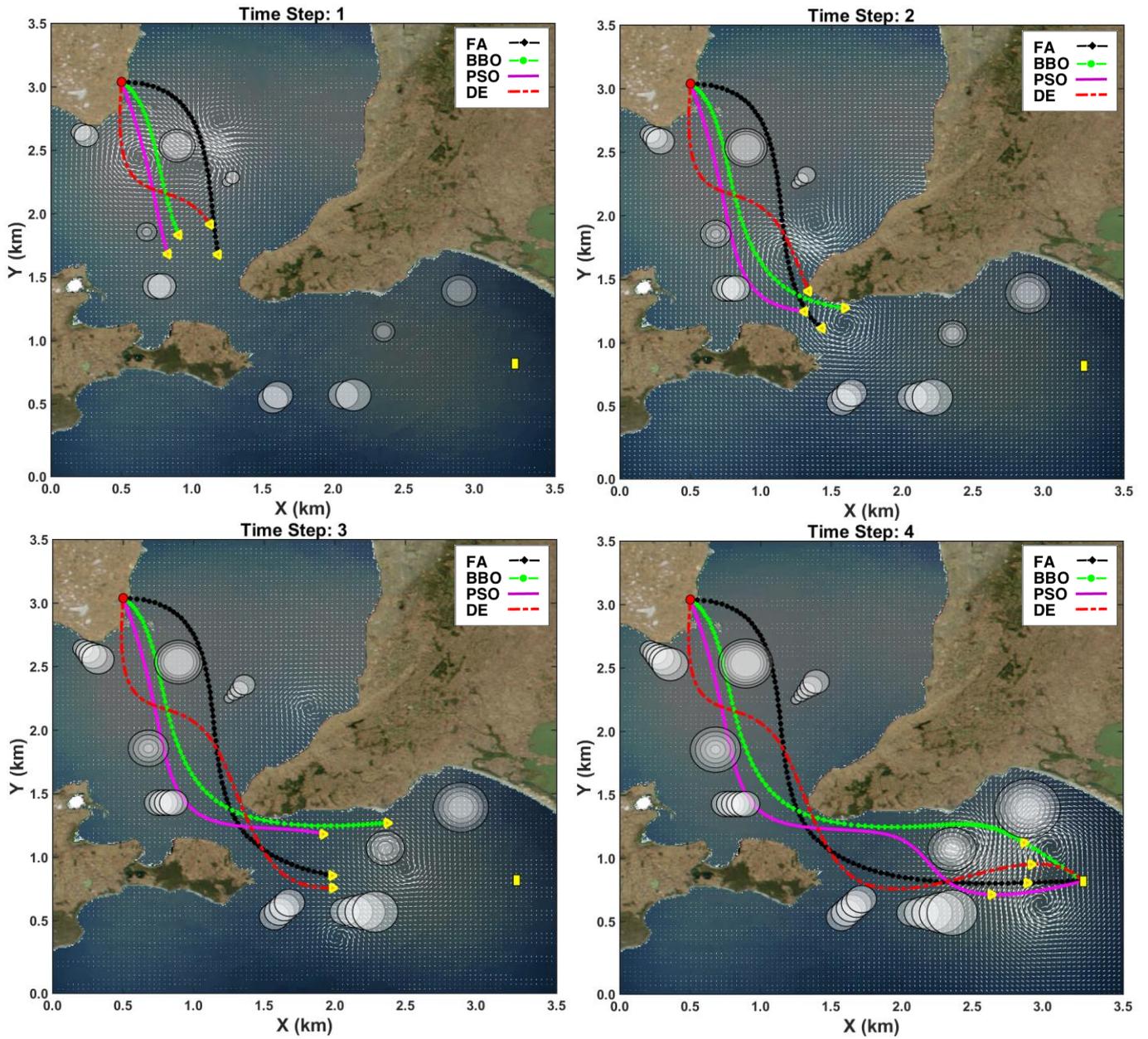

**Fig.15.** The path behaviour and re-planning process in a more complex scenario encountering different uncertain dynamic and static obstacles and current update in 4 time steps.

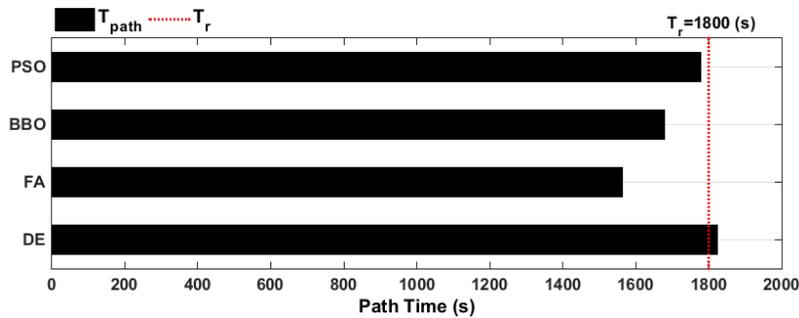

**Fig.16.** Performance of the algorithms for time optimality criterion

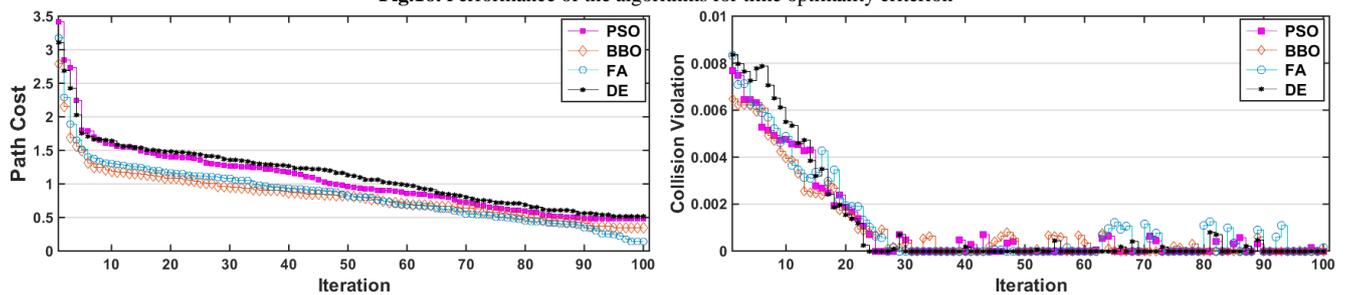

**Fig.17.** Cost and collision violation variations for FA, DE, BBO, and PSO

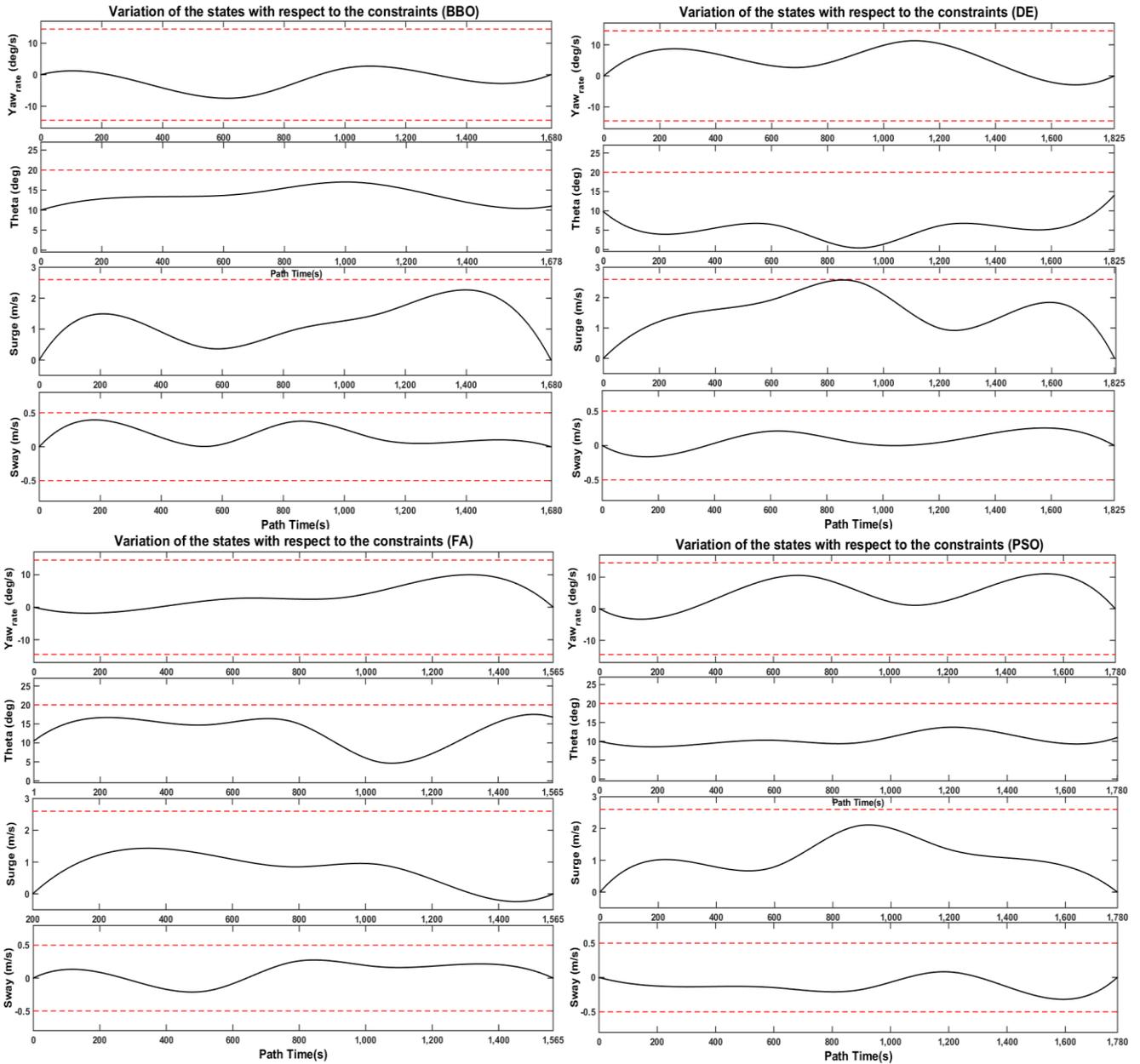

**Fig.18.** Variations of the states through the rendezvous mission with respect to the vehicular constraints (red dash line)

## 7 Conclusion

The challenging underwater rendezvous problem is addressed in this study. Using the concept of nonlinear optimal control theory, the problem is transformed into this framework and then solved using evolutionary algorithms. The rendezvous scenario is defined through a highly cluttered and variable operating field. To have an expedient adaptation with the variations of the underlying environment, a novel online path planning mechanism is developed and implemented on the follower AUV. The proposed planner is capable of refining the original path considering the update of current flows, uncertain static and moving obstacles. This refinement is not computationally expensive as there is no need to compute the path from scratch and the obtained solutions of the original path is utilized as the initial solutions for the employed methods. This leads to reduction of optimization space and accelerate the process of searching, therefore the optimal solution can be achieved in a fast manner. Applying the evolutionary path planning methods namely PSO, BBO, DE, and FA indicate that they are capable of satisfying the rendezvous problem's conditions. In details, their performance expresses that relying on the heuristic nature, they are efficient approaches to solve a complex optimal control problem. In a highly uncertain operating field, they all, are able to generate collision free path to guide the AUV toward the rendezvous place; in addition, they can remarkably satisfy the time optimality condition that is expressed by a time threshold in this study, and constraints over the corresponding states due to the physical limitations of the vehicle. These observations are confirmed by the simulation results. From comparison point of view, the utilized evolutionary methods generate slightly different solutions due to the inherent differences in their nature and mechanism. More specifically, in this study, the FA planner shows better performance in terms of making use of favorable current flow for AUV maneuverability and collision avoidance. In terms of satisfying the time optimality condition, the performance of the PSO and DE path planner is better. In summary, the simulation results confirm that the proposed online planning approach using all four meta-heuristic algorithms are efficient and fast enough in generating optimal and collision-free path encountering dynamicity of the uncertain operating field and results in leveraging the autonomy of the vehicle for having a successful mission. For future researches, internal situations and vehicles fault tolerance also will be encountered to increase vehicle total situational awareness for establishing more complete robust decision making module and implementing the proposed approach for a real field trial test.